\documentclass[10pt,twocolumn,letterpaper]{article}
\usepackage{marvosym}
\usepackage{cvpr}      % To produce the REVIEW version
\usepackage{multirow}
\usepackage[dvipsnames]{xcolor}

\definecolor{cvprblue}{rgb}{0.21,0.49,0.74}
\usepackage[pagebackref,breaklinks,colorlinks,citecolor=cvprblue]{hyperref}
\usepackage{colortbl}
\usepackage[dvipsnames]{xcolor}
\usepackage{makecell}
\usepackage{rotating}
\def\paperID{5000} % *** Enter the Paper ID here
\def\confName{CVPR}
\def\confYear{2025}

\title{EffOWT: Transfer Visual Language Models to Open-World Tracking\\ Efficiently and Effectively}

\author{
    Bingyang Wang$^{1}$, Kaer Huang$^{2}$, Bin Li$^{2}$, Yiqiang Yan$^{2}$, Lihe Zhang$^{1}$, Huchuan Lu$^{1}$, You He$^{1,3,}$\thanks{Corresponding author} \\
    {\small $^1$ Dalian University of Technology, China} {\small $^2$ Lenovo, China} {\small $^3$ Tsinghua University, China} \\
    {\tt\small 22109095@mail.dlut.edu.cn, huangke1@lenovo.com, libinm@lenovo.com,} \\
    {\tt\small yanyq@lenovo.com, zhanglihe@dlut.edu.cn, lhchuan@dlut.edu.cn, youhe\_nau@163.com}
}

\begin{document}
\maketitle 
% \begin{abstract}
%
Open-World Tracking (OWT) aims to track every object of any category, which requires the model to have strong generalization capabilities. Trackers can improve their generalization ability by leveraging Visual Language Models (VLMs). However, challenges arise with the fine-tuning strategies when VLMs are transferred to OWT: full fine-tuning results in excessive parameter and memory costs, while the zero-shot strategy leads to sub-optimal performance. To solve the problem, EffOWT is proposed for efficiently transferring VLMs to OWT. Specifically, we build a small and independent learnable side network outside the VLM backbone. By freezing the backbone and only executing backpropagation on the side network, the model's efficiency requirements can be met. In addition, EffOWT enhances the side network by proposing a hybrid structure of Transformer and CNN to improve the model's performance in the OWT field. Finally, we implement sparse interactions on the MLP, thus reducing parameter updates and memory costs significantly. Thanks to the proposed methods, EffOWT achieves an absolute gain of 5.5\% on the tracking metric OWTA for unknown categories, while only updating 1.3\% of the parameters compared to full fine-tuning, with a 36.4\% memory saving. Other metrics also demonstrate obvious improvement.
% \end{abstract}
  
\section{Introduction}
\label{sec:intro}

\begin{figure}[htp]
\includegraphics[width=1\linewidth]{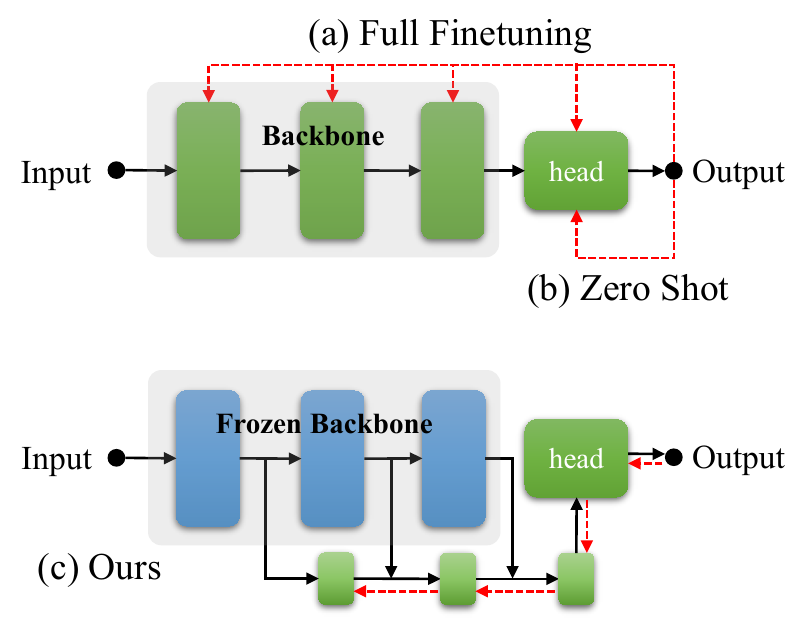}
\centering
\caption{
    A visual comparison of different fine-tuning strategies. 
    (a) The full fine-tuning does not perform any freezing operation. 
    (b) Zero-shot freezes the backbone and only fine-tunes the head. 
    (c) Our method is to build a small and independent network next to the backbone, using the intermediate features from the backbone as input. During the fine-tuning phase, the backbone is frozen. Only the parameters on the side network and head can be updated.
}
\label{fig-first}
\end{figure}

Traditional multi-object tracking (MOT) \cite{wang2020towards,pang2021quasi,sun2020transtrack} performs well on a few categories such as vehicles \cite{bdd100k} or pedestrians \cite{milan2016mot16, dendorfer2020mot20}. However, it is not suitable for some practical scenarios, such as tracking objects that have never been seen before, which is crucial for improving the safety of autonomous driving \cite{OWTB}. Traditional MOT methods are ineffective in addressing this issue because they can only track the categories predefined during the training process. Therefore, Open-World Tracking (OWT) \cite{li2023ovtrack,zheng2024nettrack} has been proposed to overcome this challenge. Different from previous MOT methods that solely track objects belonging to a limited set of classes, OWT is dedicated to tracking objects from a significantly broader range of categories, which requires stronger generalization capabilities of the model. The rapidly advancing Visual Language Models (VLMs) \cite{radford2021learning,oquab2023dinov2} can assist OWT by fine-tuning the base model on this task, effectively generalizing across thousands of categories. It has been shown that integrating VLMs into OWT can effectively enhance the model's tracking performance on the classes that are not predefined in the training set.

However, there are challenges in the fine-tuning strategy when converting VLM to OWT, such as the high parameter and memory cost during full fine-tuning \cite{li2023ovtrack}, and sub-optimal performance caused by the zero-shot strategy \cite{zheng2024nettrack}. Specifically, existing VLMs are based on the Transformer \cite{vaswani2017attention} architecture, like ViT-Huge \cite{dosovitskiy2020image} (625M) and Swin-Large \cite{liu2021swin} (197M), whose number of parameters is much larger than those of convolutional neural network (CNN) architectures, such as ResNet-50 \cite{he2016deep} (25M). Therefore, when transferring VLM to OWT, a large amount of computing resources is consumed to fine-tune the entire network completely, which reduces efficiency. In contrast, the zero-shot strategy chooses to directly freeze the backbone network of the VLM and only update the weights of the head projectors of the model. In this way, massive parameter updates and memory consumption can be avoided. However, the limited capacity of the head causes the VLM to be unable to adapt fully to downstream tasks, leading to poor performance.

Among the current OWT methods, OVTrack \cite{li2023ovtrack}, Video OWL-ViT \cite{heigold2023video}, and NetTrack \cite{zheng2024nettrack} all regard VLM as the backbone to enhance the model's generalization ability to novel categories. Specifically, OVTrack first regards CLIP \cite{radford2021learning} as the backbone network and then is fully fine-tuned on large-scale data, which suffers from the high parameter and memory consumption. Unlike OVTrack, Video OWL-ViT and NetTrack adopt the zero-shot strategy, freeze the VLM backbone, and fine-tune only the branches used for classification, regression, and correlation. Although large-scale parameter updates and memory consumption are avoided, the completely frozen VLM model fails to adapt fully to the OWT scenario and cannot achieve optimal performance. In short, neither of them manage to balance efficiency and performance. Therefore, the key lies in \textbf{how to both limit the parameters that require updating and memory costs during fine-tuning and enhance the adaptability of VLMs for OWT}.

Compared with previous methods, we construct a lightweight side network parallel to the VLM backbone to avoid the huge parameter updates and memory costs. Taking into account the characteristics of open-world tracking, the side network has been carefully designed to improve performance. Specifically: (i) We build a small and independent side network outside the backbone, which takes intermediate activations from the backbone as input and makes predictions. During the training phase, we freeze the backbone and apply backpropagation only to the side network and head. Compared with full fine-tuning, our method requires only a limited number of parameters to be updated, and also greatly reduces the memory cost during training. (ii) We construct a hybrid side network. By leveraging the locality and translational invariance of the CNN \cite{mehta2021mobilevit}, the model's ability to perceive local information is enhanced, thereby avoiding overfitting of the tracker to known classes due to long-term training. Additionally, in light of the characteristics of OWT, EffOWT proposes a multi-scale feature module to provide richer appearance information for association. (iii) To reduce the computational burden of the side network, we perform sparse interaction (SIM) on the MLP. Each token in the MLP only interacts with and fuses with tokens along its horizontal, vertical, and two diagonal directions. SIM can still capture the global receptive field, and the required parameters and memory are significantly reduced, thus further improving transfer efficiency. The main contributions are as follows:

\begin{itemize}
    \item[$\bullet$] We introduce EffOWT, which efficiently adapts the VLM for OWT by incorporating a compact and independent side network alongside the backbone. This approach updates a restricted number of parameters, while drastically reducing memory costs during training.
    
    \item[$\bullet$] We propose a Hybrid Side Network tailored for Open-World Tracking, leveraging the spatial perception capabilities of CNN to mitigate overfitting to known classes. EffOWT incorporates a multi-scale feature fusion module to enhance appearance information for ReID.

    \item[$\bullet$] We present a streamlined variant of EffOWT, which achieves a significant reduction in parameters through sparse interactions within the MLP. This optimization maintains performance while substantially reducing the computational burden on the side network.

    \item[$\bullet$] Experiments demonstrate the effectiveness and efficiency of our method. Compared with existing SOTA methods, EffOWT achieves a 5.5\% improvement in the most critical metric. During fine-tuning, EffOWT only needs to update parameters equivalent to 1.3\% of full fine-tuning, while memory requirements are also reduced by 36.4\%.
\end{itemize}

\begin{figure*}[htp]
\includegraphics[width=1\linewidth]{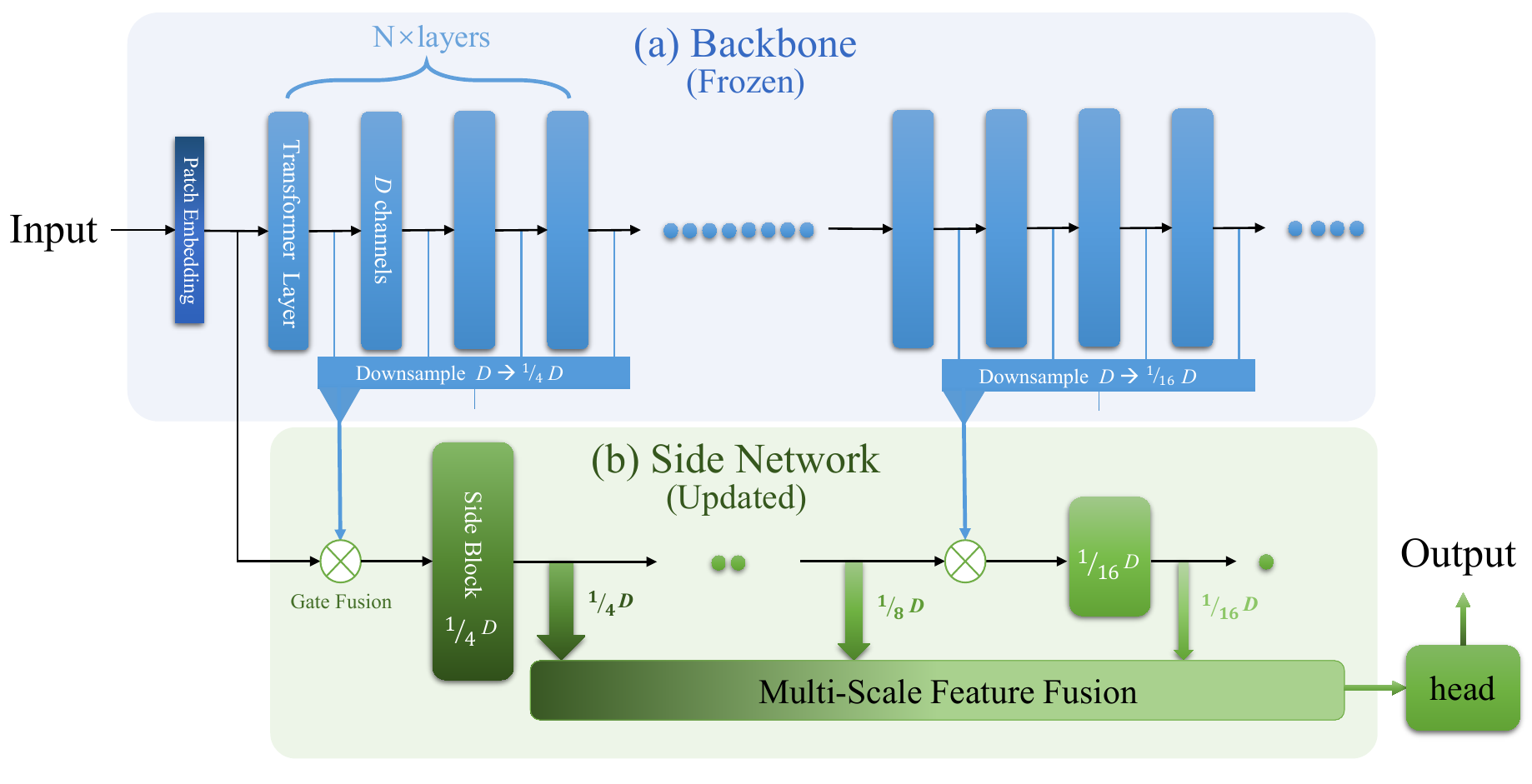}
\centering
\caption{\textbf{Overview of EffOWT.} During the fine-tuning phase, the backbone is frozen and only the parameters on the side network and head will be updated. This operation greatly reduces the parameters and memory costs required for fine-tuning. In addition, side layers are aggregated into side blocks, and CNN is introduced to form the Hybrid Side Network to avoid the model from overfitting to known classes. Finally, the multi-scale feature fusion module is proposed to provide richer appearance information to the head.}
\label{fig-overview}
\end{figure*}

\section{Related Works}
\label{sec:Related Works}
%-------------------------------------------------------------------------
% \textbf{Vision Language Model.}
\subsection{Vision Language Model}
By comprehending and analyzing image content and integrating obtained information with language data, Visual Language Models (VLM) \cite{radford2021learning, jia2021scaling} achieve a profound understanding of image content and possess strong generalization capabilities across thousands of categories. The CLIP\cite{radford2021learning} model , leveraging natural language supervision, learns visual models by combining vast amounts of unlabeled image-text pairs, emphasizing the importance of capturing visual knowledge through natural language. Compared to traditional supervised learning approaches, CLIP demonstrates significant performance improvements across multiple computer vision tasks, proving the effectiveness of natural language supervision in learning transferable visual knowledge. The ALIGN\cite{jia2021scaling} model achieves zero-shot predictions for various visual recognition tasks by learning the visual-language correspondence between numerous image-text pairs on the internet. This model simplifies the training paradigm of visual recognition tasks while reducing reliance on finely labeled data, representing a significant improvement over previous methods that relied on expensive labeled data. DINOv2\cite{oquab2023dinov2} introduces a dynamic sampling mechanism during training, which adjusts the difficulty of training samples dynamically based on the model's current performance, thus achieving more effective optimization during training. This dynamic adjustment strategy enables the model to focus on different learning objectives at different stages of training, further enhancing learning efficiency and model performance. The core of GLIP\cite{li2022grounded} lies in the introduction of a geometric perception mechanism, enabling the model to better capture spatial structures and shape information in images, thereby improving the accuracy of understanding visual content and language descriptions.  \cite{zhong2022regionclip, li2023ovtrack} focus on transferring the generalization capabilities of large-scale models to downstream tasks such as detection and tracking. The main challenge \cite{jia2022visual} lies in effectively adapting VLMs to downstream tasks while maintaining strong generalization capabilities.

% \noindent\textbf{Open-World Tracking.}
\subsection{Open-World Tracking}
Open-World Tracking\cite{OWTB} is an important research direction in the field of target tracking. It focuses on the problem of continuous tracking of unknown targets in the dynamic environment of the real world. Unlike traditional Multi-Object Tracking, the challenge faced by Open-World tracking is more complex because it requires tracking as many objects as possible, even if their categories are not predefined in the training set. This requires the model to have strong generalization. Recently, some works have introduced VLM into the field of OWT to improve the generalization of the model. OVTrack\cite{li2023ovtrack} was the first to introduce VLM into the OWT field. Specifically, it uses CLIP as a backbone to improve the model's generalization ability for targets of unknown classes. By full fine-tuning on relevant datasets, OVtrack achieves the goal of targeting many unknown classes. Accurate tracking of targets. However, the full fine-tuning strategy adopted requires updating a large number of parameters, resulting in inefficiency. Video OWL-ViT\cite{heigold2023video} and NetTrack\cite{zheng2024nettrack} also regard VLMs as backbone. Compared to full fine-tuning unilized by OVTrack, they choose to freeze the backbone directly and only fine-tune the head. But such approach prevents the model from adequately adapting to Open-World tracking scenarios, resulting in sub-optimal performance. Therefore, to solve the problem of fine-tuning and zero-shot strategies when transferring VLM to OWT, we propose EffOWT, taking into account both efficiency and effectiveness.

%-------------------------------------------------------------------------
% \noindent\textbf{Efficient Transfer.}
\subsection{Efficient Transfer}
With the rapid development in VLMs, the number of model parameters has significantly increased, leading to notable memory consumption \cite{evci2022head2toe} and computational costs \cite{zhang2020side}. To address this issue, researchers \cite{tu2023visual,xu2023side,dukler2024your,diao2023unipt} begin exploring how to improve efficiency in transfer learning while maintaining model's performance. The core of the study is to reduce the model's parameter and memory cost during the transfer learning process while maintaining or improving its performance. For example, the key of LST \cite{sung2022lst} lies in a small side network separated from backbone. This side network takes activations as inputs from intermediate layers of tbackbone and makes predictions. Backpropagation occurs solely within the side network without needing to pass through the main network, thus greatly reducing both parameter updates and memory costs. However, in practical applications, the parameters and memory usage of LST's side network during training remain relatively large, leaving considerable room for improvement. Additionally, it's crucial to note that the side network should not be viewed as a simple lightweight replica of the original network but should play a significant role in critical tasks. Therefore, to make the fine-tuning process more efficient and make the model more suitable for OWT, we propose EffOWT, which enhances from the perspectives of side network adaptation and sparse interaction on the MLP module.

\section{Methodology}
In Sec.~\ref{sec:3_1}, we introduce a Base Model by building a small and independent side network outside the backbone network, while updating limited parameters and reducing a large amount of memory cost. In Sec.~\ref{sec:3_2}, Hybrid Side Network for OWT is proposed to avoid overfitting and improve the tracking performance of the model. Finally, we propose a lightweight version of EffOWT in Sec.~\ref{sec:3_3}. By performing sparse interactions on the MLP, the burden on the side network is further reduced. The overview is shown in Fig.~\ref{fig-overview}.

\subsection{Base Model}
\label{sec:3_1}
\textbf{Open-World Tracking with Visual Language Model:} Instead of using existing MOT or OWT frameworks, we build a VLM-based tracker from scratch. Specifically: (i) Tracker construction: On the one hand, using CLIP in VLMs as the backbone, EVA is selected as the detector to achieve good generalization to unknown class targets. On the other hand, on the correlation strategy, we choose pure appearance models for matching. This is because OWT's benchmark dataset TAO-OW is shot at 30FPS and only annotated at 1FPS, and quite a few targets do not move linearly. If a motion model (such as Kalman Filter) is utilized in the inference process, performance will be reduced due to incorrect matching. (ii) Training pipeline: In the TAO-OW training set, there are very few images available for training (about 1.6w), and the images are sparsely annotated. Thus, TAO-OW cannot be used to train a passable open-world tracker. At the same time, considering that the known classes in TAO-OW overlap with those in the COCO dataset, we perform fine-tuning on the COCO dataset to train the tracker. In order to train the ReID branch on a target detection dataset such as COCO, the tracker gives the ground truths of an image according to ID numbers, and generates an additional image after data augmentation, such as flipping, based on the original image. By performing correlation matching between two sets of ground truths, the ReID branch can be fully trained. Then, we obtain an open-world tracker based on VLM, and verified the effectiveness of the method through experiments, as shown in Tab.~\ref{table_2}.

\textbf{Open-World Tracking with Side Network:} Although the above experiments have verified that either the full fine-tuning or the zero-shot strategy can achieve good performance metrics, they both still face different problems. Specifically, the parameter and memory costs are too high for full fine-tuning, resulting in low efficiency. In contrast, the zero-shot strategy of freezing the backbone and only fine-tuning the head results in suboptimal performance because the parameters on the ReID branch are too few to support excellent appearance correlation. Hence, in order to take into account both efficiency and effectiveness, the Side Network is introduced into the Open-World tracker. In detail, we build a small and independent Side Network outside the existing backbone network, which takes the intermediate features from the backbone as input and makes predictions, as shown in Fig.~\ref{fig-first}. Then, we will describe the construction, training, and limitations of the Side Network.

\textbf{Construction:} The Side Network consists of side layers and side connections. Specifically, (i) Side layers are lightweight clones of Transformer layers in the backbone. The channel dimension of each side layer is $\frac{1}{r}$ ($r$=4) of the corresponding backbone layer. Clearly, when the dimension $D(backbone_{i})$ of the $i$-th layer is 1024, then $D(side_{i})$=256. In addition, we utilize Structural Pruning on the backbone to initialize the weight of the side layers. According to the importance score, the weights ranking in the top $\frac{1}{r}$ are selected as the initial weights of the side layers to inherit the generalization of the original backbone to the greatest extent and speed up the training. (ii) Side connections are responsible for downsampling the feature map output by each Transformer layer in the backbone and fusing it with the corresponding side feature. The fused feature is regarded as the input of the next side layer. The feature fusion mechanism is performed in a dynamically weighted manner, and the specific formula is as follows:

\begin{equation} f_{i+1}^s = g_{i}*f_{i}^b+(1-g_{i})*f_{i}^s \end{equation}
where $f_{i}^b$ and $f_{i}^s$ represent the feature map of the backbone and side of the $i$-th layer, respectively, and $g$ is a learnable parameter that helps the model achieve the best feature fusion effect in an adaptive manner.

\textbf{Training process: }During training, we freeze the backbone and only update parameters on the side network and head. Then, all modules of the model will participate in the forward propagation stage, but only the side network and head perform backpropagation to improve the feature extraction and prediction capabilities of the model. Compared with full fine-tuning, the parameters that require updating are limited, and the memory cost during training is also greatly reduced. Compared with the zero-shot strategy, the side network provides assistance for VLM to fully adapt to OWT scenarios. Experiments prove that such a solution design can not only achieve efficient fine-tuning, but also maintain the generalization ability of VLM to improve the model's tracking ability of unknown categories. The conflict between efficiency and performance of previous fine-tuning strategies is basically resolved.

However, the Side Network also has certain limitations, primarily reflected in: (i) Existing side layers should not just be regarded as the lightweight version of the backbone, but rather should assist VLM in better transfer to the OWT domain. (ii) The scale of parameters to be updated and the memory cost remain relatively large, which should be further reduced to improve fine-tuning efficiency.

\subsection{Hybrid Side Network for OWT}
\label{sec:3_2}

\begin{figure}[htp]
\includegraphics[width=1\linewidth]{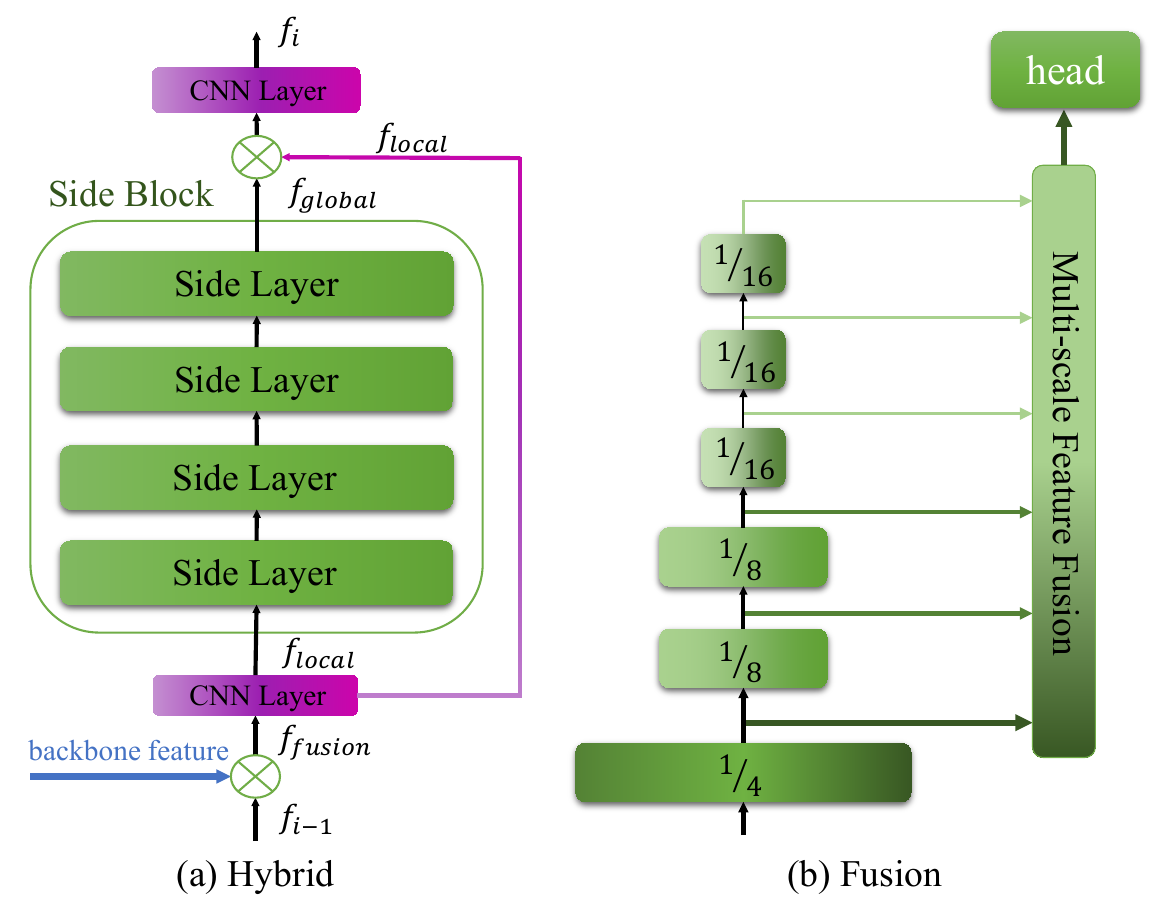}
\centering
\caption{\textbf{
An illustration of Hybrid Side Network for OWT.} (a) shows the structure of the hybrid side network and the feature processing flow. Among them, the purple block represents the CNN module. In (b), the network structure of the proposed multi-scale feature fusion module is described.}
\label{fig-LSN}
\end{figure}

In Sec.\ref{sec:3_1}, side layers are lightweight copies of Transformer layers in the backbone, but such a structure is not completely suitable for the field of open world tracking. Specifically, as demonstrated in \cite{chen2022mobile, wang2022pvt}, despite the use of relative position encoding, inductive bias is still missing in pure Transformer structures. Under such conditions, an excellent Transformer-based model requires long-term training (about 100 epochs), which is not allowed in open world tracking scenarios. As confirmed by \cite{joseph2021towards, gupta2022ow}, if a model is trained for an extended period on a dataset solely containing known classes, it will certainly incline towards the known categories, gradually losing the ability to generalize to unknown ones. Therefore, in order to avoid forgetting unknown classes due to the lack of inductive bias, we utilize the hybrid structure of CNN and Transformer to adjust the side network, as shown in Fig.\ref{fig-LSN} (a). By utilizing the locality and translation invariance of CNN, the side network's ability to perceive position information is enhanced, while we are able to take into account the advantages of the Transformer's adaptive weighting and global processing.

\begin{table}[]
\centering
\caption{
\textbf{Multi-scale side block.} The dimensions, number and parameter proportions of multi-scale side blocks.} 
\scriptsize{
\centering
\resizebox{0.76\linewidth}{!}{
\begin{tabular}{ccc}
\hline
Dimensions & Number & Parameter \\ \hline
1/4        & 1      & 59.1\%    \\
1/8        & 2      & 29.7\%    \\
1/16       & 3      & 11.2\%    \\ \hline
\end{tabular}}}
\label{tab-3_2}
\end{table}

Specifically, (i) Side block construction. Each $N$ ($N$=4) side layers are regarded as a side block, which is used to enhance the model's ability to perceive global information. Then, the CNN layers are deployed outside the side block as a module for local position information extraction. (ii) Feature processing pipeline. Firstly, the feature $f_{i-1}$ from the previous block performs weighted fusion with the intermediate features passed from the backbone to produce $f_{fusion}$. Secondly, the CNN layer performs local feature enhancement on $f_{fusion}$, resulting in $f_{local}$. Third, $f_{local}$ is input into the block for global feature processing, obtaining $f_{global}$. Finally, $f_{local}$ is fused with $f_{global}$ in a residual connection. The fused features are enhanced by a CNN layer to obtain the final output $f_{i}$, which takes into account both local and global information. (iii) In addition, in order to further reduce the parameter burden on the side network while avoiding overfitting, we perform multi-scale operations on the side network in units of side blocks. The dimensions and quantities of side blocks of different scales are shown in Tab.~\ref{tab-3_2}. In this way, by designing such a hybrid structure network, we combine the local capabilities of the CNN and the global perspective of the Transformer, effectively improving the feature extraction ability of the Side Network, and reducing its requirements for parameters and memory.

Multi-scale feature fusion module: As demonstrated in \cite{ye2022joint}, it is proven that there is a conflict between classification and ReID tasks. Specifically, the classification branch requires high-level features, while the reid tends to low-level appearance information to distinguish different objects belonging to the same category. Therefore, leveraging the multi-scale operations mentioned earlier, we integrate the feature maps of different scales produced by each side block through weighted fusion, serving as the input for the head, rather than solely relying on the feature produced by the final side block, as shown in Fig.~\ref{fig-SIM} (b). By enriching the appearance information in the features, reid receives effective support, thus enabling the side network to play a greater role in the field of Open-World Tracking.

\subsection{Sparse Interactions on MLP}
\label{sec:3_3}

\begin{figure}[htp]
\centering
\includegraphics[width=1\linewidth]{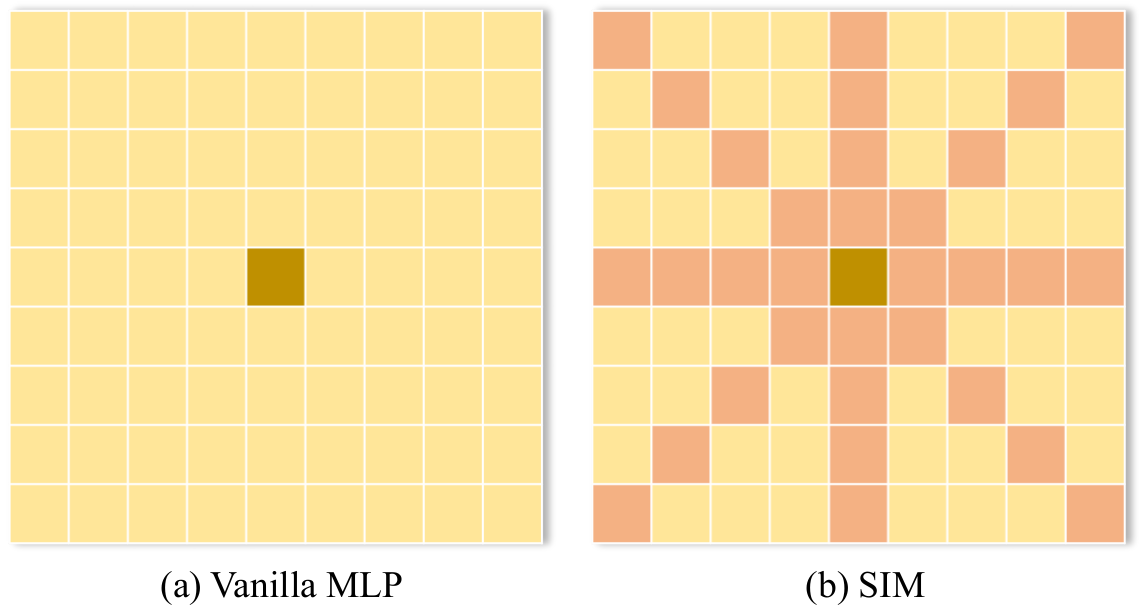}
\caption{\textbf{
A visual comparison of vanilla MLP and SIM in interactive calculations.} In a vanilla MLP, each token interacts with all other tokens. In SIM, a single token only interacts with tokens on its horizontal, vertical, and two diagonal lines.}
\label{fig-SIM}
\end{figure}

Based on the Hybrid Side Network for OWT, we provide a lightweight version to further reduce the parameter update scale and video memory requirements. According to the Transformer architecture, a single side layer consists of MHA, MLP, and Layer Norm. MHA refers to the multi-head attention mechanism, where a memory-efficient attention mechanism is used to reduce the complexity from O($n^{2}$) to O($\sqrt{n}$). The parameter amounts of MHA, MLP, and Layer Norm account for 33.3\%, 66.6\%, and 0.1\%, respectively. In other words, the main obstacle restricting further lightweighting of network parameters lies in the MLP.

\begin{table*}[]
\centering
\caption{
\textbf{Results comparison on the TAO-OW val.} 
The comparison is shown in terms of OWT related metrics including Open-World Tracking Accuracy (OWTA), Detection Recall (D.Re), and Association Accuracy (A.Acc) for both unknown and known classes.} 
\label{tab:freq}
\resizebox{\linewidth}{!}{
\begin{tabular}{lcccccc}
\midrule[0.5pt]
& \multicolumn{3}{c}{{\color[HTML]{000000} Unknown}} & \multicolumn{3}{c}{{\color[HTML]{000000} Known}}   \\
Methods               & OWTA↑                & D.Re↑                & A.Acc↑      & OWTA↑   & D.Re↑                & A.Acc↑               \\ 
\midrule
\multicolumn{7}{l}{\textcolor[rgb]{0.5, 0.5, 0.5}{\textit{a) Closed-World}}} \\
\midrule
Tracktor\cite{sridhar2019tracktor}              & 23.7                & 56.3                & 10.4       & 57.8    & 80.5                & 42.4                \\
OC-SORT\cite{cao2023observation}               & 31.6                & 37.8                & 27.1       & 48.7    & 69.0                & 34.4                \\
SORT\cite{bewley2016simple}                  & 35.3                & 45.8                & 30.5       & 47.3    & 69.2                & 33.4                \\
SORT-TAO\cite{dave2020tao}              & 39.9                & 68.8                & 24.1       & 54.2    & 74.0                & 40.6                \\
ByteTrack\cite{zhang2022bytetrack}             & 40.9                & 40.5                & 42.5       & 63.3    & 71.3                & 56.4                \\
\midrule
\multicolumn{7}{l}{\textcolor[rgb]{0.5, 0.5, 0.5}{\textit{b) Open-World}}} \\
\midrule
OWTB\cite{OWTB}                 & 39.2                & 46.9                & 34.5       & 60.2    & 77.2                & 47.4                \\
NetTrack\cite{zheng2024nettrack}              & 43.7                & 58.7                & 33.2       & 62.7    & 77.4                & 51.0                \\
Video OWL-ViT\cite{heigold2023video}         & 47.3                & 62.3                & 37.2       & 56.6    & 73.2                & 44.6                \\
OVTrack\cite{li2023ovtrack}               & 48.6                & 58.0                & 42.0       & 46.0    & 52.3                & 42.6                \\
SimOWT\cite{wang2023simple}                & 50.4                & 63.3                & 41.5       & 62.9    & 78.7                & 50.9                \\
EffOWT & \textbf{56.1(+5.7)} & \textbf{71.1(+7.8)} & \textbf{44.8(+3.3)} & \textbf{68.8(+5.9)} & \textbf{86.6(+7.9)} & \textbf{55.0(+4.1)} \\
\bottomrule
\end{tabular}}
\end{table*}

\begin{table*}[]
\centering
\caption{
\textbf{Results comparison on fine-tuning strategies.} 
The comparison is shown in terms of Open-World Tracking related metrics and Efficient Transfer metrics including Parameter and Memory.} 
\label{tab:freq}
\resizebox{1\linewidth}{!}{
\begin{tabular}{ccccccccc}
% \toprule
\midrule[0.5pt]
& & & \multicolumn{3}{c}{{\color[HTML]{000000} Unknown}} & \multicolumn{3}{c}{{\color[HTML]{000000} Known}}   \\
Methods  & Parameter↓ &  Memory↓    & OWTA↑     & D.Re↑  & A.Acc↑     & OWTA↑   & D.Re↑    & A.Acc↑ \\ 
\midrule
Zero-shot &  -   & 10.3G       & 47.2    & 67.1       & 33.7     & 59.6     & 85.3          & 41.9    \\
Full Finetuning &    349.2M          & 20.9G        & 47.6    & 64.7       & 35.6      & 63.5    & 86.2  & 47.0    \\
EffOWT              & 7.0M(2.0\%)       & 14.1G(67.5\%)    & \textbf{56.1(+8.5)}              & \textbf{71.1(+6.4)}       & \textbf{44.8(+9.2)}           & \textbf{68.8(+5.3)}           & \textbf{86.6(+0.4)}          & \textbf{55.0(+8.0)}    \\
EffOWT\dag    & \textbf{4.5M(1.3\%)}       & \textbf{13.3G(63.6\%)}   & 55.9(+8.3)     & 70.3(+5.6)       & 44.9(+9.3)  & 68.2(+4.7)  & 86.4(+0.2)          & 54.1(+7.1)    \\
\bottomrule
\end{tabular}}
\label{table_2}
\end{table*}

During propagation, in a vanilla MLP, each token interacts with all other tokens, as shown in Fig.~\ref{fig-SIM} (a). The purpose of such interaction operations is to enable the MLP to obtain a global perception. However, the drawback is that it results in an excessively large number of parameters being utilized. To achieve further lightweighting, we perform Sparse Interaction in the MLP. SIM is divided into two modules: the interaction module and the fusion module. Specifically, the interaction module allows a single token to interact only with tokens located along the horizontal, vertical, and two diagonal lines of its position, thus capturing characteristic information in each direction to the greatest extent. The fusion module is responsible for weighted fusion of four identity maps containing the output of each path and the original features. Although such operations are sparse, they still allow SIM to obtain the global receptive field while significantly reducing the number of parameters in the MLP.

In the multi-scale operation (Sec.\ref{tab-3_2}), the side block with a dimension of $\frac{1}{4}$ of the backbone accounts for 59.1\% of the parameters, becoming the main obstacle to further lightweighting of the Side Network. Therefore, we deploy the SIM module on the side block of this scale while avoiding its introduction in smaller-scale side blocks. This is due to the fact that, as spatial features are progressively down-sampled, the reduced number of tokens significantly diminishes the interaction capability, which has been shown to lead to substantial performance degradation (as demonstrated in \cite{cao2023strip}). To validate this design choice, we conducted comparative experiments and observed that deploying SIM in smaller-scale side blocks resulted in notable performance drops (e.g., in accuracy or inference efficiency). In contrast, deploying SIM at the $\frac{1}{4}$ scale not only effectively reduces the parameter count but also retains the global perceptual ability, yielding significant performance improvements. This design achieves an optimal trade-off between network lightweighting and performance, offering a promising approach for further parameter reduction.

\section{Experiments}
In this section, we first introduce the experimental setup of the Open-World tracking task in Sec.~\ref{sec:4_1}, which includes the evaluation metrics, datasets and implementations details of the experiments involved. In Sec.~\ref{sec:4_2}, our approach is compared with SOTA method on the tracking metrics in the Open-World. Furthermore, comparisons with full fine-tuning and zero-shot strategies verify the effectiveness of our EffOWT. Then, we conducted extensive ablation experiments in Sec.~\ref{sec:4_3}. Finally, we visually compare the tracking performance between SOTA and EffOWT in Sec.~\ref{sec:4_4}.

\subsection{Experimental Setup}
\label{sec:4_1}
\noindent\textbf{Evaluation metrics.} Indicators are divided into two categories: (i) Open-World tracking-related indicators, which are utilized to evaluate the model's tracking, detection, and correlation capabilities for both known and unknown classes. (ii) Efficient Transfer indicators, which are utilized to describe the number of parameters required for updating and the memory cost during the fine-tuning stage.

Among the Open-World tracking-related indicators, following the configuration of methods such as OWTB \cite{OWTB} and Video OWL-ViT \cite{heigold2023video}, the main indicator, Open-World Tracking Accuracy (OWTA), evaluates the tracking performance of the model for both known and unknown classes. Furthermore, OWTA is calculated from two indicators: Detection Recall (D.Re) and Association Accuracy (A.Acc). Specifically, OWTA is obtained by taking the geometric mean of D.Re and A.Acc, where the positioning threshold $\alpha$ is utilized for evaluation. OWTA is also calculated by integrating over the range of $\alpha$ values.

% \begin{equation}
% \textrm{$A.Acc_{\alpha}$} = {1 \over |\textrm{$TP_{\alpha}$}|} \sum_{c \in \textrm{$TP_{\alpha}$}}{\textrm{$TPA_{\alpha}$}(c) \over {\textrm{$TPA_{\alpha}$}(c) + \textrm{$FPA_{\alpha}$}(c)} + \textrm{$FNA_{\alpha}$}(c)}
% \end{equation}
% \begin{equation}
% \textrm{$D.Re_{\alpha}$} = {|\textrm{$TP_{\alpha}$}| \over |\textrm{$TP_{\alpha}$}| + |\textrm{$FN_{\alpha}$}|}, \textrm{$OWTA_{\alpha}$} = \sqrt{D.Re_{\alpha} \cdot A.Acc_{\alpha}}
% \end{equation}

It is worth noting that the choice of recall as the detection metric instead of precision or mAP is related to the task setting of open-world tracking. Open-World Tracking aims to detect and track as many objects as possible, regardless of whether their categories are present in the training set. However, it is impractical for the dataset to label all foreground classes in an image. In fact, the TAO-OW \cite{dave2020tao} dataset used for the OWT task is sparsely annotated. Therefore, D.Re, which does not penalize false positives, is selected. In summary, OWTA, D.Re, and A.Acc are key indicators in the field of open-world tracking, responsible for evaluating the tracking, detection, and correlation performance of the model on known and unknown categories, respectively.

On the other hand, as for the efficient transfer-related indicators, parameter refers to the parameter scale required for updating the backbone and side network during training, and memory refers to the average peak memory usage by the model per iteration during training.

\begin{table*}[]
\centering
\caption{
\textbf{Ablation studies on each component of our main method.} 
} 
\label{tab:freq}
\resizebox{1\linewidth}{!}{
\begin{tabular}{ccccccccccc}
% \toprule
\midrule[0.5pt]
\multicolumn{3}{c}{{\color[HTML]{000000} Methods}} & & & \multicolumn{3}{c}{{\color[HTML]{000000} Unknown}} & \multicolumn{3}{c}{{\color[HTML]{000000} Known}}   \\
Base Model & HSN & SIM   & Parameter↓ &  Memory↓    & OWTA↑     & D.Re↑  & A.Acc↑     & OWTA↑   & D.Re↑    & A.Acc↑ \\ 
\midrule
        $\checkmark$    &      &  & 25.8M(7.4\%)       & 14.6G(69.9\%)      & 52.7              & 66.4       & 42.5            & 67.1           & 86.7          & 52.1    \\
 $\checkmark$   &     $\checkmark$      &      & 7.0M(2.0\%)       & 14.1G(67.5\%)    & 56.1              & 71.1       & 44.8            & 68.8           & 86.6          & 55.0    \\
$\checkmark$ & $\checkmark$ &  $\checkmark$   & 4.5M(1.3\%)       & 13.3G(63.6\%)   & 55.9     & 70.3       & 44.9  & 68.2 &86.4          & 54.1    \\
\bottomrule
\end{tabular}}
\label{table_3}
\end{table*}

\begin{table*}[]
\centering
\caption{
\textbf{Ablation studies on each component of HSN.} 
} 
\label{tab:freq}
\resizebox{1\linewidth}{!}{
\begin{tabular}{ccccccccccc}
% \toprule
\midrule[0.5pt]
\multicolumn{3}{c}{{\color[HTML]{000000} Methods}}  & & & \multicolumn{3}{c}{{\color[HTML]{000000} Unknown}} & \multicolumn{3}{c}{{\color[HTML]{000000} Known}}   \\
Base Model & Hybrid & Fusion   & Parameter↓ &  Memory↓    & OWTA↑     & D.Re↑  & A.Acc↑     & OWTA↑   & D.Re↑    & A.Acc↑ \\ 
\midrule
        $\checkmark$    &      &  & 25.8M(7.4\%)       & 14.6G(69.9\%)      & 52.7              & 66.4       & 42.5            & 67.1           & 86.7          & 52.1    \\
$\checkmark$ & $\checkmark$ &    & 6.4M(1.8\%)       & 13.9G(66.5\%)   & 55.0              & 70.2       & 43.6            & 68.8           & 87.1          & 54.6    \\
 $\checkmark$   &     $\checkmark$      &  $\checkmark$     & 7.0M(2.0\%)        & 14.1G(67.5\%)    & 56.1              & 71.1       & 44.8            & 68.8           & 86.6          & 55.0    \\
\bottomrule
\end{tabular}}
\label{table_4}
\end{table*}

\noindent\textbf{Datasets.} COCO \cite{coco} and TAO-OW \cite{dave2020tao} datasets are utilized for training and validation, respectively. COCO consists of 80 categories, covering many frequently occurring categories in daily life. To train the ReID module on COCO, the tracker obtains the ground truth labels using the ID numbers of the images and generates additional images by applying data augmentation methods such as flipping and random cropping on the original one. The embedding module of the model is trained by matching between two sets of ground truth labels. Through this process, COCO can be regarded as an Open-World Tracking dataset to train a tracker with certain generalization capabilities.

TAO-OW contains 833 categories, covering a wide range of categories and scenarios. Some categories are rare in daily life, such as paddles and parachutes, which pose a challenge to the generalization ability of the model. According to the settings of OWTB, TAO-OW is divided into known classes and unknown classes. Specifically, the known categories overlap with those included in COCO, such as people, vehicles, etc. Unknown categories refer to those that do not appear in COCO, such as juicers and aquatic plants.

We choose COCO as the training set for the following reasons: (i) Although the TAO-OW dataset is shot at 30FPS, it is only annotated at 1FPS, and the images are sparsely annotated, resulting in a limited number of images and corresponding annotations available for training (approximately 1.6w images and 5.4w annotations). In other words, the TAO-OW dataset is insufficient to train a competent open-world tracker. (ii) COCO contains 80 categories, which overlap with the known categories in the TAO-OW setting. Training on COCO prevents category leakage and ensures that the model is exposed only to known categories during training. In the inference stage, since TAO-OW's test set annotation and evaluation server are not yet public, we evaluate our model on the validation set containing 988 videos provided by TAO-OW. The performance of the model is evaluated on both known and unknown classes. Both the training and inference stages are performed using a class-agnostic approach.

\begin{figure*}[htp]
\centering
\includegraphics[width=0.98\linewidth]{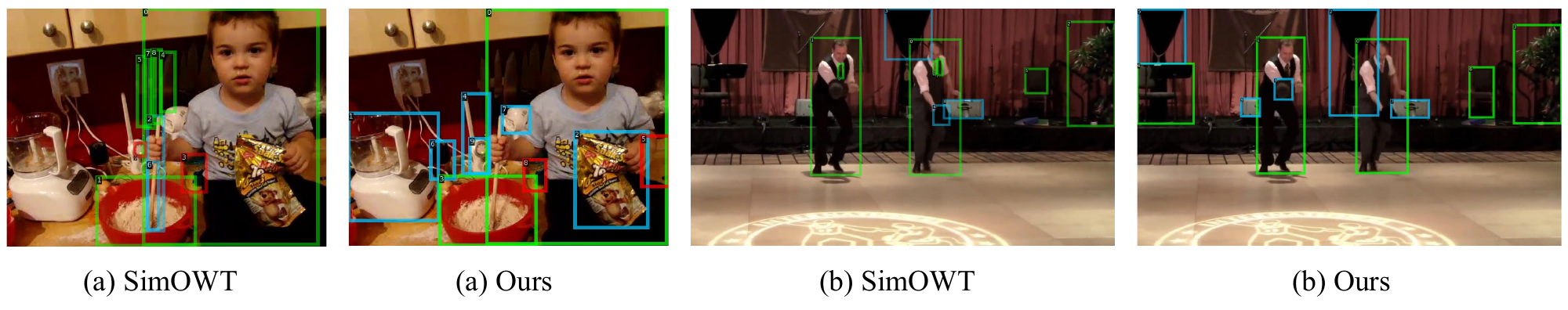} 
\caption{\textbf{Visual comparison between SOTA method SimOWT and our EffOWT.} The visual comparison is conducted under that all these models provide top 10 predictions. Specifically, we use the green box to denote the prediction box which overlaps well with a known object. The boxes colored blue are used to represent the potential and unknown objects. And the other boxes are colored red to indicate some meaningless or inaccurate predictions.}
\label{fig-vis} 
\end{figure*}
 
\noindent\textbf{Implementation details.} (i) In the field of Open-World Tracking, there are challenges inherent in full fine-tuning. The reasons are as follows: Firstly, due to the lack of annotations during training, unknown class instances are treated as background. As training progresses, the model's performance on unknown categories gradually deteriorates. Secondly, because full fine-tuning involves updating the VLM backbone, the model's generalization ability is compromised. The ultimate result is that the full fine-tuning strategy yields suboptimal results on OWT tasks, as shown in Tab.~\ref{table_2}. (ii) EffOWT consists of the Base Model and Hybrid Side Network for OWT. In addition, based on EffOWT, we proposed a more lightweight version, called EffOWT\dag, by using Sparse Interactions on MLP.

\subsection{State-of-the-art Comparison} 
\label{sec:4_2} 
The comparison between our method and current Open-World tracking algorithms is shown in Tab.~\ref{table_1}, which is divided into two parts: Closed-World and Open-World. Closed-World algorithms are trained on TAO-OW, using both unknown and known class data. In contrast, Open-World algorithms do not use any unknown-class data in TAO-OW during training. Experiments show that, due to the effectiveness of our approach, EffOWT achieves 56.1\% OWTA, 71.1\% D.Re, and 44.8\% A.Acc on unknown classes. Compared with the current best OWT method, SimOWT, our EffOWT achieves an absolute gain of 5.7\% on unknown class OWTA, an improvement of 7.8\% on D.Re, and 3.3\% on A.Acc. In the known categories, OWTA, D.Re, and A.Acc improved by 5.9\%, 7.9\%, and 4.1\%, respectively. Compared with the state-of-the-art (SOTA) methods, the significant improvement across all indicators fully demonstrates the effectiveness of EffOWT. Compared with other open-world tracking algorithms that utilize VLMs (such as NetTrack, Video OWL-ViT, and OVTrack), EffOWT also significantly exceeds them in relevant indicators, strongly proving the rationality of our method.

Tab.~\ref{table_2} shows the comparison of our EffOWT with the full fine-tuning and zero-shot strategies on efficient transfer (parameters and memory) and open-world tracking-related metrics. In the experiment, the zero-shot strategy froze the backbone during fine-tuning and only updated the parameters on the head. Full fine-tuning allows the backbone and head to be jointly trained. According to the experimental results, during the training phase, EffOWT only requires 1.3\% of the learnable parameter updates and 63.6\% of the memory cost compared to full fine-tuning. Meanwhile, compared with full fine-tuning, EffOWT achieves absolute gains of 8.3\%, 5.6\%, and 9.3\% on OWTA, D.Re, and A.Acc, respectively, for unknown classes. As for known classes, EffOWT also outperforms full fine-tuning. Additionally, EffOWT demonstrates remarkable superiority over the zero-shot strategy across all metrics for both known and unknown categories. Moreover, EffOWT achieves a reduction of parameter costs by 0.7\% and memory requirements by 3.9\%, rendering the model more lightweight. The experimental results demonstrate the contribution of our approach to the OWT community.

\subsection{Ablation Study}
\label{sec:4_3}
We design the ablation study for the main method and HSN to show the contribution of each component.

\noindent\textbf{Main method.} In Tab.~\ref{table_3}, we study three methods: Base Model, HSN, and SIM, to show how they influence the final performance of our method. Specifically: (i) By deploying the Hybrid Side Network for OWT module on the Base Model, there is a reduction of 5.4\% in the parameters requiring updates during training, along with a 2.4\% decrease in memory costs. Additionally, concerning OWT metrics, compared to the baseline, the model exhibited a 3.4\% enhancement in OWTA for the unknown category, a 4.7\% improvement in D.Re, and a gain of 2.3\% in A.Acc. This illustrates HSN's ability to reduce model weight while also enhancing model generalization. (ii) Introducing SIM into the model, based on the Base Model and HSN, leads to notable reductions in parameter requirements of 0.7\% and memory demand of 3.9\%. The experimental findings confirm the effectiveness of SIM in further optimizing the model for lightweight applications.

\noindent\textbf{HSN.} In Tab.~\ref{table_4}, we focus on the ablation experiments of the HSN module. The introduction of the Hybrid structure leads to a reduction of 5.6\% in the parameters requiring updates compared to the baseline and a 3.4\% decrease in memory costs. Additionally, there is a 2.3\% increase in OWTA for the unknown category, a 3.8\% improvement in D.Re, and a 1.1\% increase in A.Acc. Moreover, the model's performance on known classes is also enhanced. (ii) The Fusion in Table 5 denotes the multi-scale feature fusion module. By incorporating the Fusion module into the side network, although there is a slight increase of 0.2\% in parameter count and 1.0\% in memory cost, the model exhibits a 2.4\% increase in OWTA for the unknown class, a 3.4\% improvement in D.Re, and a gain of 1.6\% in A.Acc. Moreover, there is some enhancement observed in the indicators of known categories. Experimental results highlight the effect of HSN in reducing parameter and memory costs while enhancing model performance.

\subsection{Visualization} 
\label{sec:4_4} 
The visual comparison depicted in Fig.~\ref{fig-vis}, illustrates the disparity between SimOWT and our method. EffOWT shows superior generalization capabilities compared to SimOWT, particularly on unknown categories, while maintaining commendable performance on known categories. For instance, EffOWT excels in accurately tracking a broader array of unknown class targets, including household appliances, curtains. Overall, EffOWT is better suited for OWT than SimOWT.
\section{Conclusion}
We propose EffOWT, which extends visual language models to open-world tracking. Compared with previous methods, the tracker addresses the inefficiency or sub-optimal performance of fine-tuning strategies during transfer. By building a small and independent side network outside the backbone network to extend the basic model, parameter updates and memory costs are significantly reduced. Next, we introduce a hybrid side network to enhance tracking performance while mitigating the risk of overfitting. Finally, implementing sparse interactions on the MLP further reduces the computational burden on the side network. Experiments demonstrate that EffOWT greatly reduces parameter and memory costs during the fine-tuning phase. Furthermore, EffOWT shows excellent performance on TAO-OW, surpassing state-of-the-art (SOTA) methods on both known and unknown categories, which demonstrates the effectiveness of our method. We hope this work will establish a solid baseline for future research in the field of OWT.

{
    \small
    \bibliographystyle{ieeenat_fullname}
    \bibliography{ref_tnnls}
}

% \bibliography{references} 
% \input{main.bbl} 
% \input{X_suppl}

\end{document}